\documentclass[conference]{IEEEtran}
\IEEEoverridecommandlockouts
\usepackage{cite}
\usepackage{amsmath,amssymb,amsfonts}
\usepackage{algorithmic}
\usepackage{graphicx}
\usepackage{multirow}
\usepackage{textcomp}
\usepackage{xcolor}
\def\BibTeX{{\rm B\kern-.05em{\sc i\kern-.025em b}\kern-.08em
    T\kern-.1667em\lower.7ex\hbox{E}\kern-.125emX}}
\begin{document}

\title{Information Elevation Network for Fast Online Action Detection}

\author{\IEEEauthorblockN{Sunah Min\IEEEauthorrefmark{1}\IEEEauthorrefmark{2} and 
Jinyoung Moon\IEEEauthorrefmark{1}\IEEEauthorrefmark{2}}
\IEEEauthorblockA{\IEEEauthorrefmark{1}University of Science and Technology, Daejoen, Republic of Korea}
\IEEEauthorblockA{\IEEEauthorrefmark{2}Electronics and Telecommunications Research Institute, Daejoen, Republic of Korea}
\thanks{This work was supported by Institute of Information \& Communications Technology Planning \& Evaluation (IITP) grant funded by the Korean government (MSIT) (No. 2020-0-00004, Development of Previsional Intelligence based on Long-term Visual Memory Network and No. 2014-3-00123, Development of High Performance Visual BigData Discovery Platform for Large-Scale Realtime Data Analysis).}
\thanks{Corresponding author: Jinyoung Moon (email: jymoon@etri.re.kr).}}

\maketitle

\begin{abstract}
Online action detection (OAD) is a task that receives video segments within a streaming video as inputs and identifies ongoing actions within them. It is important to retain past information associated with a current action. However, long short-term memory (LSTM), a popular recurrent unit for modeling temporal information from videos, accumulates past information from the previous hidden and cell states and the extracted visual features at each timestep without considering the relationships between the past and current information. Consequently, the forget gate of the original LSTM can lose the accumulated information relevant to the current action because it determines which information to forget without considering the current action. We introduce a novel information elevation unit (IEU) that lifts up and accumulate the past information  relevant to the current action in order to model the past information that is especially relevant to the current action. To the best of our knowledge, our IEN is the first attempt that considers the computational overhead for the practical use of OAD. Through ablation studies, we design an efficient and effective OAD network using IEUs, called an information elevation network (IEN). Our IEN uses visual features extracted by a fast action recognition network taking only RGB frames because extracting optical flows requires heavy computation overhead. On two OAD benchmark datasets, THUMOS-14 and TVSeries, our IEN outperforms state-of-the-art OAD methods using only RGB frames. Furthermore, on the THUMOS-14 dataset, our IEN outperforms the state-of-the-art OAD methods using two-stream features based on RGB frames and optical flows.
\end{abstract}


\maketitle

\section{Introduction}
\label{sec:introduction}
\begin{figure*}[htbp]
    \centering
    \includegraphics[scale=0.5]{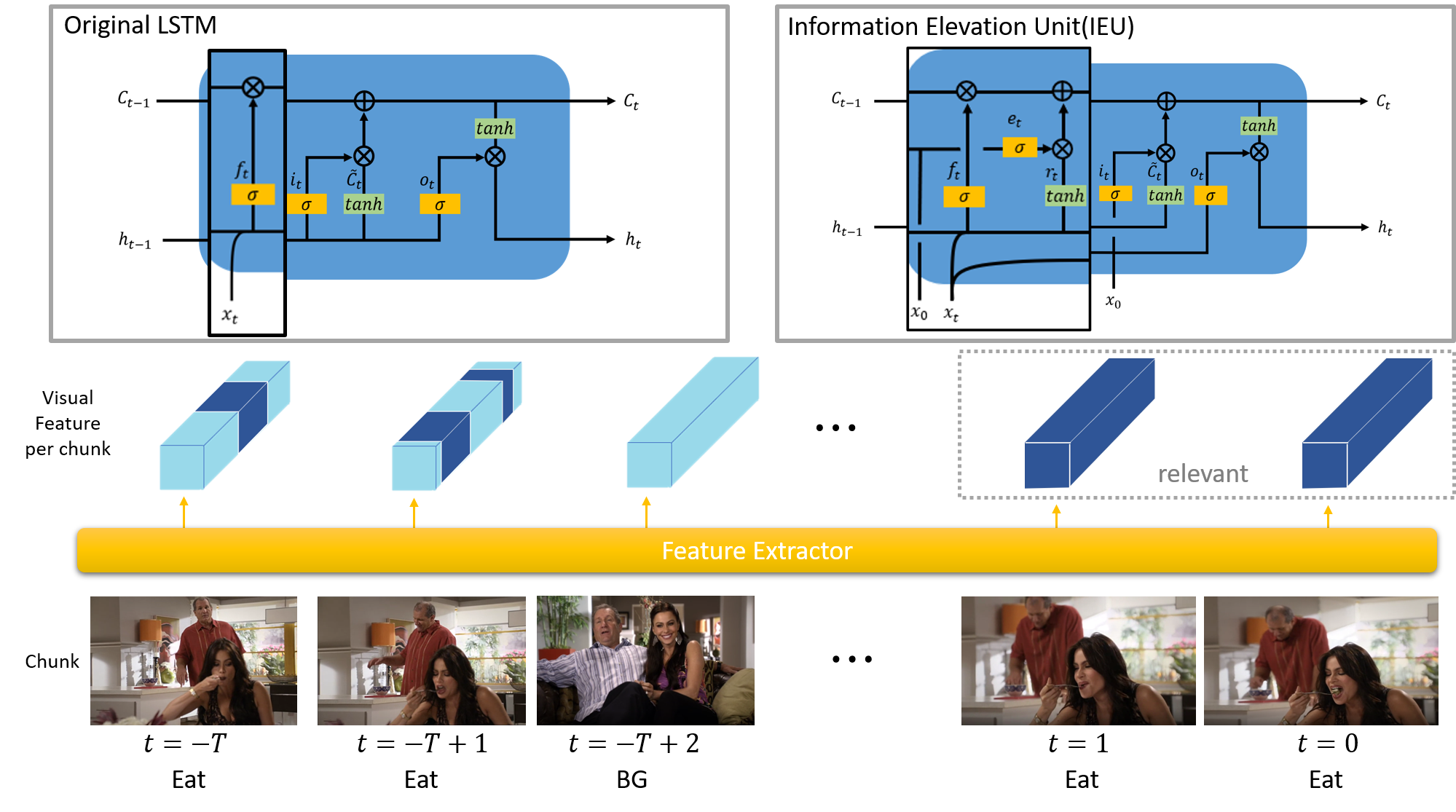}
    \caption{Comparison between the original LSTM and our information elevation unit (IEU). In this video segment, past information at times $-T$ and $-T+1$ is related to the current action. However, when processing the information at time $-T+1$, the LSTM considers only past information from the previous hidden and cell states and at $-T+1$ timestep. In the LSTM, there is a risk of removing accumulated information relevant to current action at the forget gate and accumulating information at the timestep that is irrelevant to current action at the input and output gates. Therefore, the proposed IEU takes the current information together with the past information as inputs and adds an elevation gate to maintain and accumulate the past information relevant the current action.}
    \label{fig:introduction}
\end{figure*}
In recent times, the video surveillance industry has experienced dramatically growth with the proliferation of CCTV cameras \cite{b1}. Accordingly, the demand for analyzing surveillance videos to detect actions as soon as they happen has also increased. For analyzing a large volume of streaming videos as inputs, efficient and effective action detection is essential.
\par 
Action detection (AD) studies are largely divided into offline and online AD methods. Offline AD methods receive a fully observed untrimmed video including multiple action instances and background; they then localizes temporal intervals for the action instances and identify their action classes. Most AD methods have been dealt with offline temporal AD. However, offline AD approaches process an entire input video, and then a time delay occurs for detecting all action instances. In contrast, online action detection (OAD) methods take a partially observed video segment as an input and recognize the current action of the latest frame within the input video segment. A video segment consists of a fixed number of chunks, each of which is composed of a fixed number of consecutive frames. The methods obtain a visual feature for each chunk, which is extracted from a pre-trained action recognition network. To sum up, an OAD method takes a video segment as an input and predicts the action class of the last chunk by using visual features extracted from all chunks within the video segment.
\par
Several OAD methods have been proposed. Reinforced encoder-decoder (RED) \cite{b2} and temporal recurrent network (TRN) \cite{b3} predict the future information in advance and use the predicted future information to detect the ongoing action. RED and TRN assume that all chunks constituting a video segment are equally related to the current action. However, not all chunks of a video segment are related to the current action. Therefore, information discrimination network (IDN) \cite{b4} and temporal filtering network (TFN) \cite{b5} detect the current action for a video segment by selecting or emphasizing the visual information from chunks related to the current action, considering the relationship between information from every chunk and the last chunk. Thus, it is necessary to take into account the relevance of the information to the current action in designing OAD methods.\par
Most neural networks that process time-series data such as videos use long short-term memory (LSTM) \cite{b6}. However, LSTM takes temporal inputs at every timestep from the previous hidden and cell states (i.e., $h_{t-1}$ and $C_{t-1}$) and a visual feature (i.e., $x_t$) and does not consider the current information at $t=0$ (i.e., $x_0$). In particular, the forget gate of LSTM can lose the past information by using only visual information at each timestep (i.e., $x_t$) and accumulated information from the previous hidden state (i.e., $h_{t-1}$). In Fig. \ref{fig:introduction}, even though the information related to the current action, \emph{Eat}, exists in $t=-T$ and $t=-T+1$, the relevant information may not be maintained at the next $t={-T+2}$ because LSTM, when processing visual information at $t={-T+2}$, which are relevant to \emph{background}, regards only those accumulated information from the previous hidden and cell states as unnecessary. \par
To this end, we propose an information elevation unit (IEU) for extended LSTM with an additional gate for OAD. To maintain the information related to the current action, the IEU has the information elevation gate that lifts up the past information relevant to the current action to the cell state. Specifically, the IEU adds the past information from the previous hidden state (i.e., $h_{t-1}$) as well as the visual feature at $t$ timestep (i.e., $x_t$) multiplied by the output from the elevation gate considering the relationship between the accumulated past and current information (i.e., $h_{t-1}$ and $x_0$). Through this, the IEU can reinforce the forgotten past information relevant to the current action.\par
In addition, we designed our IEN based on the proposed IEU by considering time efficiency as well. Existing OAD methods detect a current action using visual features from RGBs as well as optical flows. However, obtaining optical flows requires a large amount of computation time and computing resources, and using optical flows for OAD obstructs the use of OAD for real-world online applications. Therefore, we exclude optical flows and extract visual features from only RGB frames.\par
To show the effectiveness and efficiency of our IEN consisting of IEUs, we conduct experiments using two OAD benchmark datasets, THUMOS-14 and TVSeries. Taking the visual features from only RGB frames, the IEN outperforms state-of-the-art OAD methods using only RGB frames with a per-frame mAP of 60.4\% and mcAP of 81.4\% in THUMOS-14 and TVSeries, respectively. In particular, the IEN also outperforms the state-of-the-art methods using visual features from both RGB and optical flow frames in THUMOS-14.\par
Our contribution is summarized as follows.
\begin{itemize}
\item We propose a novel RNN unit for OAD, called IEU, which extends an LSTM unit by adding an additional information elevation gate and taking the current information together with the conventional temporal inputs of LSTM at each timestep as inputs to maintain and accumulate the past information relevant to the current action.
\item For practical use of OAD for real-world online applications, we adopt a fast action recognition network using only RGB frames as a feature extractor.
\item For the THUMOS-14 and TVSeries datasets, the proposed IEN achieves the best OAD performance compared to state-of-the-art OAD methods using visual features from only RGB frames. In addition, on the THUMOS-14 dataset, our IEN outperforms the state-of-the-art OAD methods using visual features from RGB frames as well as optical flows.
\end{itemize}
\section{Related work}\label{sec2}
\subsection{Offline action detection}
Offline AD methods take an untrimmed video as an input and detect one or more actions within the video \cite{b18}. The detection results include the temporal intervals of all detected actions with their start and end times and their action classes. Shou \emph{et al.} \cite{b7} introduced proposal, classification, and localization networks based on three-dimensional convolution neural networks (3D CNNs) for offline action detection. After generating some video segments of different lengths, the part related to action instances and the part related to background are identified through the proposal network. The classification network recognizes what the action is in the candidate interval. Finally, the localization network estimates the temporal overlap between GT and the candidate interval. Lin \emph{et al.} \cite{b8} generated proposals by using the estimated start and end scores for action instances rather than generating a proposal of a fixed length. However, adopting offline AD methods for processing surveillance videos have a limitation: there is a delay in obtaining detection results because of processing time for detecting all action instances within the input video. Some AD methods have also been proposed for detecting specific actions in surveillance videos \cite{b19,b20}.

\subsection{Online action detection}
OAD was introduced by Geest \emph{et al.} \cite{b9}. OAD is the task of predicting what the current action is for a given video segment. Given an input video segment, OAD methods predict the probability distribution for the current action located to the last chunk of the video segment, among $K$ action classes and background. RED \cite{b2} is based on the encoder-decoder structure and predicts the current and future actions by using the past information obtained from the encoder. The cell of a temporal recurrent network, which is proposed by Xu \emph{et al.} \cite{b3}, generates future information by using the past information and predicts the current action using the past information and the generated future information. In order to predict the current action, both these methods consider all timestamps with the same weight. To improve this, the information discrimination network (IDN) \cite{b4} and temporal filtering network (TFN) \cite{b5} processed the information of each timestamp based on the relevance of the current action. The information discrimination unit (IDU), which was proposed in the IDN, is a modified unit of the gated recurrent unit (GRU) \cite{b10}. It distinguishes the information related to the ongoing action among the information at time $t$ and transfers it to the next hidden state. TFN extracts visual features for all chunks within an input video segment and concatenates the extracted features for a visual embedding. Taking this concatenated visual embedding as an input, TFN predicts a relevance vector indicating the degree of relevance to the current chunk through the three kinds of filtering modules based on convolution layers. Then, TFN obtains a modified visual embedding by multiplying the relevance vector to the original visual embedding by considering the relationships between every chunk at time $t$ and the current chunk. Finally, TFN outputs the current action using the modified visual embedding.

\subsection{Action recognition}
For a given well-trimmed video containing a single action instance, action recognition (AR) methods predict the probability distribution for N action classes. AR is a task that has been studied for a long time, and networks based on convolutional neural networks (CNNs) have shown good performances. Proposed AR networks have been used as feature extractors for various video-related tasks including offline and online temporal action detection. Convolutional three-dimensional network (C3D) \cite{b11} and inflated three-dimensional convolutional network (I3D) \cite{b12} are representative AR models used as visual feature extractors for videos. C3D extracts temporal information by extending 2D CNN to 3D CNN in order to solve the limitation that extracting features using the existing 2D CNN does not extract temporal information well. I3D extracts spatiotemporal visual features by extending the 2D CNN of the existing two-stream network to 3D CNN. Both the AR models show good performance, but they require heavy computation loads. To overcome the lack of consideration of the diversity of scales in temporal domain, Wan \emph{et al.} \cite{b13} proposed an AR method based on two-stream CNNs, for extracting long-term spatiotemporal features using stacked RGB images and short-term spatiotemporal features using optical flows from two adjacent RGB frames.

\subsection{Fast action recognition}
Most existing AR methods use optical flow as inputs or are based on 3D CNN to extract temporal information. However, a 3D CNN requires substantial computation, and it also takes a large amount of time and memory to extract optical flows between consecutive RGB frames in an input video. Therefore, fast AR methods that efficiently extract temporal information using few parameters have been proposed. The temporal shift module (TSM) \cite{b14} extracts visual features for each frame chunk of a video and then concatenates these features for the video to create a feature map. Then, the channel of this feature map is divided into three parts, shifted to +1, -1, and 0, based on the time axis. The TSM creates a new feature map through 2D CNN. As TSM uses only a simple shift operation, temporal information is obtained without additional parameters. Zhang \emph{et al.} \cite{b15} proposed a novel motion cue called persistence of appearance (PA) using RGB frames. They calculate PA between two adjacent frames only accumulating pixel-wise differences in feature space, instead of using exhaustive patch-wise search of all the possible motion vectors as optical flows do.
\section{Proposed Method}\label{sec3}
\begin{figure}[htbp]
    \centering
    \includegraphics[scale=0.35]{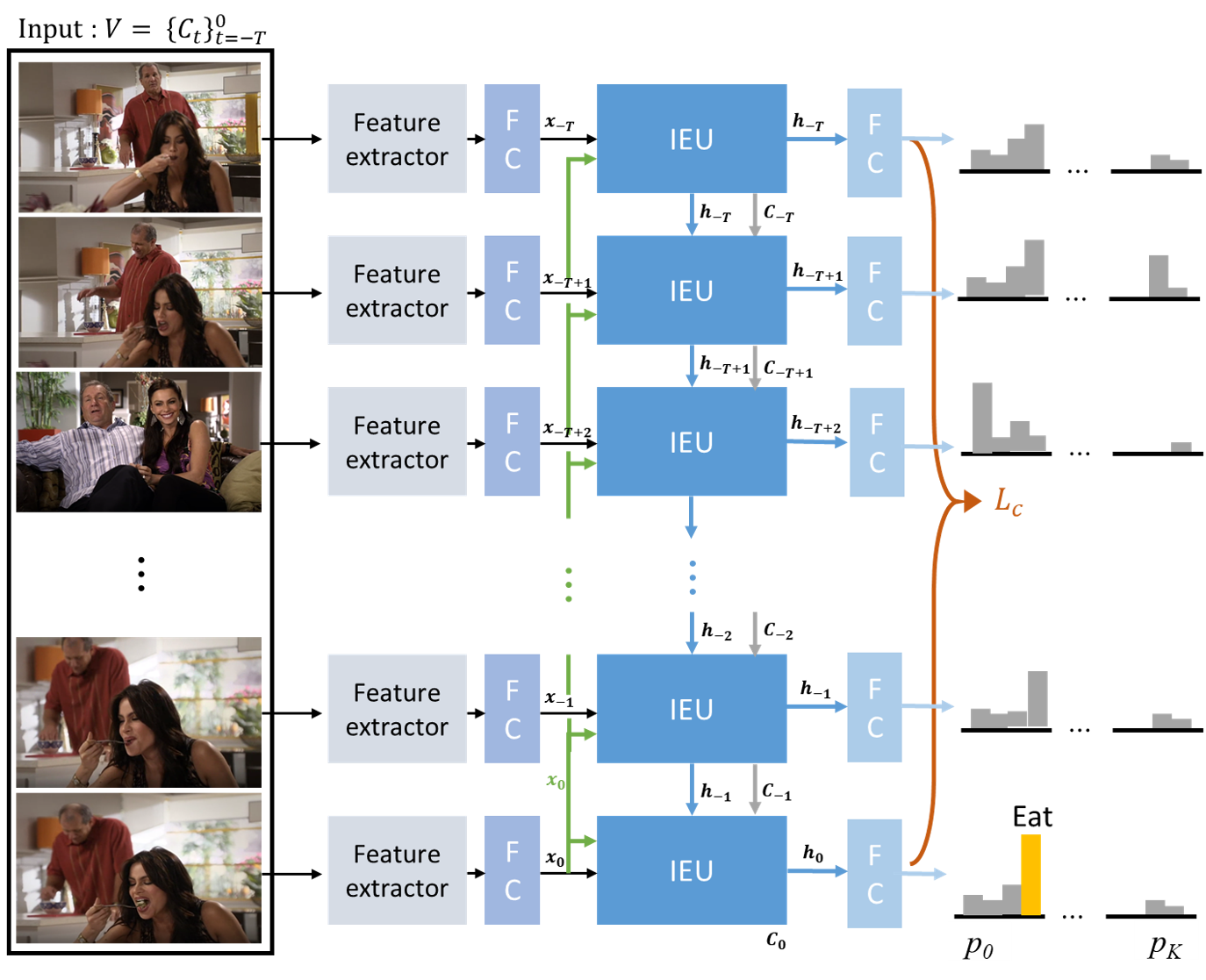}
    \caption{Architecture of the IEN. Taking a video segment consisting of $T+1$ chunks $V=\{c_t\}_{t=-T}^0$ as an input, IEN obtains each embedding vector by extracting visual features for each chunk and embedding the extracted features. The embedding vector is generated for each chunk and put into the IEU. At this time, the feature at $t=0$ representing the current action is entered together with the feature at each time $t$. Loss, $L_c$, and probabilities are calculated for the outputs of all IEUs, and the probabilities for the K action classes and the background in the last chunk are used to determine the current action.}
    \label{fig:Network}
\end{figure}

Figure \ref{fig:Network} shows the overall architecture of our IEN. For a given video segment, the IEN extracts a visual feature for each chunk, converts the visual feature into a visual embedding, and feeds it into each IEU. Each IEU takes past information from previous hidden and cell states related to a visual feature at each timestep $t$ and the current information from the current chunk at time $t=0$. Using each hidden state at each timestep $t$, IEN predicts the probability distribution for each chunk and returns the probability distribution for the current chunk within the video segment. In this section, we explain component modules including early embedding, IEU, and classification modules, in detail.

\subsection{Problem definition}
The goal is to detect the ongoing action for a given video segment. In OAD for a streaming video, we split the video into video segments with a fixed number of chunks. A chunk refers to a sequence of consecutive frames, denoted as $c=\{I_n\}_{n=1}^N$ of a set of N consecutive frames, where ${I_n}$ indicates the $n_{th}$ frame. Taking a video segment, i.e., a sequence of chunks, as an input, an OAD model predicts the current action, which means the action of the last chunk of the video segment. To formulate the OAD problem, we follow the same setting as in previous methods \cite{b2, b3}. Given a streaming video $V=\{c_t\}_{t=-T}^0$ including current (i.e., at $t=0$) and T past chunks (from $t=-T$ to $t=1$) as inputs, an OAD model outputs the probability distribution $p_0=\{p_{0, k}\}_{k=0}^K$ of the current action at chunk $c_0$ over $K$ action classes and background.
\subsection{Early embedding module}
Given a video segment $V$ as input, the early embedding module generates visual embedding for each chunk. We used TSM \cite{b14}, a fast AR model, for efficiently extracting spatiotemporal features using only RGB frames. We denote the extracted features for the video segment as $V' = \{c_0' \cdots c_T'\}\in\mathbb{R}^{(T+1) \times d_v}$. The extracted feature per each chunk contains spatiotemporal information. We feed the extracted feature into a fully connected layer to generate the embedded visual feature $x_t =$ ELU$(W_c \cdot c_t') \in \mathbb{R}^{d_e}$, where $W_c \in \mathbb{R}^{d_v \times d_e}$ is a weight matrix, through the ELU activation function \cite{b21}.

\subsection{Information Elevation Unit (IEU)}
\begin{figure}[htbp]
    \centering
    \includegraphics[scale=0.38]{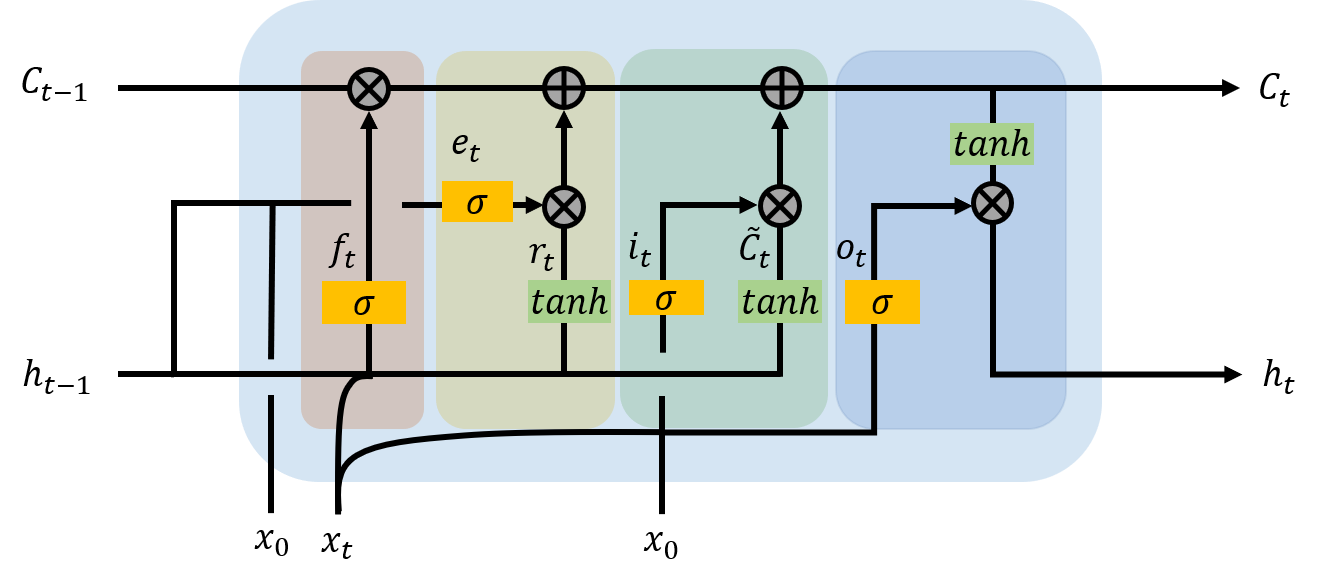}
    \caption{Structure of the information elevation unit (IEU). The IEU is an extended LSTM by adding an additional elevation gate and taking additional input, current information $x_0$. The IEU’s forget gate (red box) is the same as the original LSTM and input and output gate (green box) is similar except the input $x_0$. The elevation gate (yellow box) is newly added. Merging lines implies the addition operation between vectors.}
    \label{fig:IEU}
\end{figure}

We propose IEU, a novel RNN unit, using the current information, which is the most important for the OAD task. The IEU extends the original LSTM unit. First, IEU additionally takes the current visual embedding to determine which information is relevant to the current action at the modified input and output gates and newly-added elevation gate. Second, the IEU adds a new gate called an elevation gate in order to lift up accumulated past information from the previous cell state relevant to the current action, which can be forgotten at the forget gate. The elevation gate takes $x_0\in \mathbb{R}^{d_e}$ and the previous hidden state $h_{t-1} \in \mathbb{R}^{d_h}$ to determine which information from the previous hidden state is relevant to the current action. Figure. \ref{fig:IEU} illustrates the structure of the IEU for OAD. The equations related to all the gates of the IEU are expressed as follows.
\begin{equation}
    f_t = \sigma({W_f}\cdot({h_{t-1} + x_t })),
    \label{eq:forget_gate}
\end{equation}
\begin{equation}
    e_t = \sigma({W_e}\cdot({h_{t-1}} + x_0)),
    \label{eq:elevation_gate}
\end{equation}
\begin{equation}
    r_t = tanh({W_r}\cdot({h_{t-1} + x_t})),
    \label{eq:elevation_relevance_gate}
\end{equation}
\begin{equation}
    i_t = \sigma({W_i}\cdot({x_t + x_0})),
    \label{eq:input_gate}
\end{equation}
\begin{equation}
    \tilde{C_t} = tanh({W_c}\cdot({h_{t-1} + x_t})),
    \label{eq:cell_gate}
\end{equation}
\begin{equation}
    C_t = (C_{t-1} \times f_t) + (r_t \times e_t) + (\tilde{C_t} \times i_t),
    \label{eq:new_cell_state}
\end{equation}
\begin{equation}
    o_t = \sigma({W_o}\cdot(x_t + x_0))),
    \label{eq:output_gate}
\end{equation}
\begin{equation}
    h_t = o_t \times tanh(C_{t}),
    \label{eq:hiddenstate_gate}
\end{equation}
 where $W_f, W_e, W_r, W_c \in\mathbb{R}^{(d_h + d_e) \times d_h}$, $W_i, W_o  \in\mathbb{R}^{(d_e + d_e) \times  d_h}$ are learnable parameters.

\subsubsection{Forget and Elevation Gates}
As in (\ref{eq:forget_gate}), the forget gate of IEU is identical to that of the original LSTM. The forget gate determines which accumulated information from the cell state should be forgotten without using the current information at time $t$. However, the current information is directly related to the output of an OAD model. As a result, even though the accumulated past information from the previous cell states can be related to the visual embedding at $t=0$, there is a risk that the past information can be forgotten if it has less of a relationship with the previous hidden state at the forget gate. To overcome this limitation, the IEU locates an elevation gate next to the forget gate. The current information (i.e., $x_0$) as well as the past information from the previous hidden state (i.e., $h_{t-1}$) instead of the past information at timestep $t$ (i.e., $x_t$) are taken as input. First, as shown in (\ref{eq:elevation_gate}), the elevation gate determines which past information is reinforced through the sigmoid function with the previous hidden state and the current information. The IEU obtains the output of $r_t$ by taking the past information from the previous hidden state and the visual embedding at time $t$ in as inputs in  (\ref{eq:elevation_relevance_gate}). To compensate for the forgotten information relevant to the current action, the IEU adds the result of multiplying $e_t$ and $r_t$ to the cell state in (\ref{eq:new_cell_state}). 

\subsubsection{Input and Output Gates}
As in (\ref{eq:output_gate}) and (\ref{eq:input_gate}), the input and output gates of IEU are modified by taking different inputs, the past information at timestep $t$ (i.e., $x_t$ and the current information (i.e., $x_0$) instead of the previous hidden state (i.e., $h_{t-1}$) to control the two gates according to the current information. The input and output gates determines which past information at timestep $t$ is relevant to the current information. The output of the input gate affects the next cell state and that of the output gate affects the next hidden state.

\subsection{Classification Module}
In the classification module, we predict a sequence of probability distributions for all the $T+1$ chunks within a video segment over K+1 classes as $p$ as in (\ref{eq:classification module_1}) by feeding a sequence of $T+1$ hidden states from $-T$ to $0$ as:
\begin{equation}
    p = \ [p_{-T,k}, p_{-T+1,k}, ..., p_{0,k}]_{k=0}^{K} , 
    \label{eq:classification module_1}
\end{equation}
\begin{equation}
    h = \ [h_{-T} | h_{-T+1} | ... | h_{0}], 
    \label{eq:classification module_2}
\end{equation}
\begin{equation}
    p = softmax({W_{cls}},{h}),
    \label{eq:classification module_3}
\end{equation}
where | is the concatenation operation and $W_{cls,t}\in\mathbb{R}^{{d_h} \times {d_{cls}}}$ is a trainable matrix. 
To train our IEN, we define a classification loss for a sequence of the $T+1$ chunks over K+1 classes by employing the cross-entropy loss as: 
\begin{equation}
    L_{cls} = - \sum_{t=-T}^{0} \sum_{k=0}^{K} y_{t,k}\log(p_{t,k}),
    \label{eq:loss func}
\end{equation}
where $y_{t,k}$ is the ground-truth label for the $t_{th}$ timestep. The cross-entropy loss is applied to each chunk so that the hidden state from all chunks can contain the information for all action sequences. Finally, our IEN returns $p_0$ = $[p_{0,k}]_{k=0}^{K}$ its final output.
\section{Experiments}\label{sec4}
We conducted experiments with two OAD benchmark datasets, THUMOS-14 and TVSeries. First, this section gives an overview of these datasets. Second, we explain the evaluation metrics used to evaluate OAD performance and describe the experimental settings for implementing the proposed IEN. Third, we compared the performances of state-of-the-art methods to our IEN on both the OAD datasets. Finally, we evaluate three versions of LSTM variants to show the efficiency and effectiveness of our IEU through an ablation study.
\subsection{Datasets}
\subsubsection{THUMOS-14}
THUMOS-14 \cite{b22} is a dataset initially publicized for a competition for offline action detection and localization. THUMOS-14 collects videos from YouTube. This dataset was divided into 20 action classes related to sports such as diving and tennis swing. As the training set of THUMOS-14 consists of well-trimmed videos, its validation set was used for training and its test set was used for AD evaluation. Specifically, 200 validation videos were used for training and 213 test videos were used for testing in the experiment.
\subsubsection{TVSeries}
TVSeries \cite{b9} is a realistic dataset consisting of 27 episodes from six famous TV series. Each video contains a single episode whose length is approximately 20 minutes or 40 minutes. The 27 videos are divided into 13, 7, and 7 for training, validation, and test set, respectively. A total of 6,231 action instances over 30 classes appear in this dataset. As its videos are collected from TV series, the dataset includes actions with large variability, with the appearance of several actors and actions occurring simultaneously.

\subsection{Evaluation metrics}
Following the evaluation protocol in \cite{b2, b3, b4, b5, b9}, we used the mean average precision (mAP) and mean calibrated average precision (mcAP) at the frame-level for THUMOS-14 and TVSeries, respectively, as the output of OAD models are the probability distribution over K+1 classes for the current frame (i.e., chunk).   

\subsubsection{Mean average precision}
Based on the probabilities at all frames predicted by an OAD model over $K$ action classes, all frames were first sorted into descending order. The precision of a class at cut-off $i$ is calculated as:
\begin{equation}
    Prec(i) = \frac{TP(i)}{TP(i)+FP(i)},
    \label{eq:Precision}
\end{equation}
is the number of true-positive frames and FP(i) is the number of false-positive frames at cut-off $i$. The average precision of an action class is then defined as:
\begin{equation}
    AP =  \frac{\sum_{n=1}^{N}Prec(n) \cdot I(n)}{P},
    \label{eq:AP}
\end{equation}
where N is the number of frames used for evaluation, P is the total number of positive frames, and I(n) is 1 when the $n_{th}$ frame is true positive and 0 otherwise. The per-frame mean AP (mAP) over K classes is calculated as:
\begin{equation}
    mAP = \frac{\sum_{k=1}^{K}AP_k}{K}
    \label{eq:mAP}
\end{equation}

\subsubsection{Mean calibrated average precision}
For the TVSeries dataset, the per-frame mean calibrated average precision (mcAP) is used as a performance evaluation metric. The per-frame mAP has the disadvantage that it is sensitive to ratio changes between positive and negative frames. To compensate for this limitation, De Geest \emph{et al.} \cite{b9} proposed mcAP as an improved performance evaluation metric. In contrast to mAP, mcAP is based on the calibrated average precision (cAP) instead of AP. The cAP is defined as:
\begin{equation}
    cPrec(i) = \frac{TP(i)}{TP(i)+\frac{FP(i)}{w}},
    \label{eq:cPrec}
\end{equation}
\begin{equation}
    cAP = \frac{\sum_{n=1}^N cPrec(n) \cdot I(n)}{P},
    \label{eq:cAP}
\end{equation}
where $w$ represents the ratio of positive to negative frames. The per-frame mcAP over K classes is defined as:
\begin{equation}
    mcAP = \frac{\sum_{k=1}^{K}cAP_k}{K},
    \label{eq:mcAP}
\end{equation}

\subsection{Experimental Setting}
We set the frame rate of all videos at 24 fps. Each chunk consisted of 32 frames and each video segment consisted of eight chunks. We extracted visual features for each chunk. As a feature extractor, we used TSM \cite{b14} pre-trained with kinetics and extract features from the last global average pooling layer. In the early embedding module, the dimensions of the extracted features and the embedded features, $d_v$ and $d_e$, were 2,048 and 512, respectively. In the IEU, we set the hidden units to 512. We used Adam as the optimizer, set the learning rate to 0.00001, and set the batch size to 64.

\subsection{Performance comparison}
In this section, we compare our IEN to existing OAD state-of-the-art methods on the two benchmark datasets, THUMOS-14 \cite{b22} and TVSeries \cite{b9}. The OAD models are divided into those using RGB features only (noted as \emph{RGB}) and those using two-stream features using RGB as well as optical flow frames (noted as \emph{RGB+Flow}). As a feature extractor for \emph{RGB+Flow} input, RED \cite{b2}, TRN \cite{b3}, IDN \cite{b4}, and TFN \cite{b5} use the two-stream (TS) CNN \cite{b17} to extract the same TS features using RGB frames for the appearance and optical flows for the motion, as described in TRN \cite{b3} in detail. For \emph{RGB} input, RED \cite{b2} and TRN \cite{b3} use the same VGG-16 features \cite{b16}, as described in TRN \cite{b3} in detail, and IDN \cite{b4} and TFN \cite{b5} use only the appearance part of the TS features. For RGB features, our IEN use TSM \cite{b14} with only RGB frames, which was pre-trained using the Kinetics dataset \cite{b12}. 

\begin{table}[tbh!]
\centering
\caption{Performance comparison on THUMOS-14 \cite{b22}. }
\centering
\begin{tabular}{lllr}
\hline
Input & Method & Feature Extractor & mAP (\%) \\ \hline
\multirow{2}{*}{RGB}                                & TFN \cite{b5}             & TS-RGB \cite{b17}                        & 45.5              \\ \cline{2-4}
                               & \textbf{Ours}   & TSM-RGB \cite{b14}              & \textbf{60.4}  \\ \hline
\multirow{4}{*}{RGB+Flow}     & RED \cite{b2}             & \multirow{4}{*}{TS \cite{b17}}             & 45.3              \\
                               & TRN \cite{b3}             &                            & 47.2              \\ 
                               & IDN \cite{b4}             &                         & 50.0              \\
                               & TFN \cite{b5}             &                            & 55.7              \\ \hline
\end{tabular}
\label{thumos_performance}
\end{table}

Table \ref{thumos_performance} reports the results on THUMOS-14 \cite{b22}. For \emph{RGB} input, our IEN achieves a per-frmae mAP of 60.4\%, which is significantly superior to TFN with its per-frame mAP of 45.5\%. Furthermore, our IEN taking \emph{RGB} input outperforms all the state-of-the-art methods that take \emph{RGB+Flow} as input.  

\begin{table}[tbh!]
\centering
\caption{Performance comparison on TVSeries \cite{b9}.}
\centering
\begin{tabular}{lllr}
\hline
Input & Method & Feature Extractor & mcAP (\%) \\ \hline
\multirow{5}{*}{RGB}      & RED\cite{b2}           & \multirow{2}{*}{VGG\cite{b16}}    & 71.2     \\
                      & TRN\cite{b3}           &                   & 75.4     \\ \cline{2-4}
                      & IDN\cite{b4}           & \multirow{2}{*}{TS-RGB\cite{b17}}               & 76.6     \\
                      & TFN\cite{b5}           &                   & 79.0     \\ \cline{2-4}
                      & \textbf{Ours} & TSM-RGB\cite{b14}      & \textbf{81.4}     \\ \hline
\multirow{4}{*}{RGB+Flow} & RED\cite{b2}      & \multirow{4}{*}{TS\cite{b17}}            & 79.2     \\
                      & TRN\cite{b3}           &                   & 83.7     \\ 
                      & IDN\cite{b4}           &                & 84.7     \\
                      & TFN\cite{b5}           &                   & 85.0     \\ \hline
\end{tabular}
\label{tvseries_performance}
\end{table}

In Table \ref{tvseries_performance}, we summarize the results with TVSeries \cite{b9} .
For \emph{RGB} input, our IEN achieves a per-frame mcAP of 81.3\% and outperforms all the state-of-the-arts methods. However, our IEN that takes \emph{RGB} input outperforms the RED among the state-of-the-art methods that take \emph{RGB+Flow} as input. This result shows that OAD for TVSeries requires both motion and appearance information, compared to OAD for THUMOS-14 \cite{b22}. In TVSeries \cite{b9}, its actions that belong to different action classes occur in a similar environment.

Table \ref{thumos_perf_each_class} and \ref{tvseries_perf_each_class} present the per-frame AP and cAP values of all action classes on THUMOS-14 \cite{b22} and TVSeries \cite{b9}, respectively, sorted in a descending order. In Table \ref{thumos_perf_each_class}, actions either related to interactions between people and small objects or with small movements achieve relatively low detection performance because capturing their visual characteristics is difficult. In Table \ref{tvseries_perf_each_class}, actions related to small gestures or those that occur very quickly (such as \emph{point}) achieve relatively low detection performance.

\begin{table}[tbh!]
\caption{Performance of our IEN for each class on THUMOS-14 \cite{b22}.}
\centering
\begin{tabular}{lrlr}
\hline
Class & AP (\%) & Class & AP (\%) \\ \hline
BasketballDunk  & 85.9  &   HighJump        & 59.0  \\
LongJump        & 79.0  &   ThrowDiscuss       & 58.7   \\
HammerThrow     & 78.6  &   GolfSwing      & 58.4    \\
CliffDiving     & 75.2  &   CricketBowling & 56.2     \\
CleanAndJerk    & 72.3  &   SoccerPenalty      & 52.9\\
Diving          & 71.5  &   VolleyballSpiking  & 50.4    \\
PoleVault       & 70.9  &   BaseballPitch  & 46.5    \\
TennisSwing     & 67.5  &   Billiards      & 44.8    \\
Shotput         & 62.5  &   FrisbeeCatch   & 27.5   \\
JavelinThrow    & 62.4  &   CricketShot    & 26.7   \\  \hline
\multicolumn{2}{l}{mAP (\%)} & \multicolumn{2}{r}{60.4}    \\ \hline
\end{tabular}
\label{thumos_perf_each_class}
\end{table}

\begin{table}[tbh!]
\centering
\caption{Performance of our IEN for each class on TVSeries \cite{b9}.}
\centering
\begin{tabular}{lrlr}
\hline
Class & cAP (\%) & Class & cAP (\%) \\ \hline
Punch              & 98.4   &   Open door          & 81.1    \\
Fire weapon       & 95.2    &   Go down stairway   & 80.6   \\
Drive car          & 94.8   &   Sit down           & 79.8    \\
Get in/out of car & 94.6    &   Answer phone      & 79.3   \\
Kiss              & 93.6    &   Stand up           & 78.4   \\
Run                & 90.1   &   Write              & 75.9   \\
Use computer       & 90.0    &  Close door         & 75.7    \\
Drink              & 88.9   &   Pick something up  & 75.7   \\
Fall/trip         & 87.5   &   Smoke              & 73.0   \\
Go up stairway    & 87.8    &   Throw something   & 73.0    \\
Clap              & 87.0   &    Give something     & 72.4  \\
Wave              & 85.7    &   Undress           & 68.9    \\
Read               & 85.5    &  Hang up phone     & 64.4     \\
Eat               & 83.0    &   Point              & 63.4   \\
Pour              & 81.5   &   Dress up          & 56.0    \\ \hline
\multicolumn{2}{l}{mcAP (\%)} & \multicolumn{2}{r}{81.4}    \\ \hline
\end{tabular}
\label{tvseries_perf_each_class}
\end{table}
\subsection{Ablation studies}

\begin{figure*}[t]
\centering
\includegraphics[width=\textwidth]{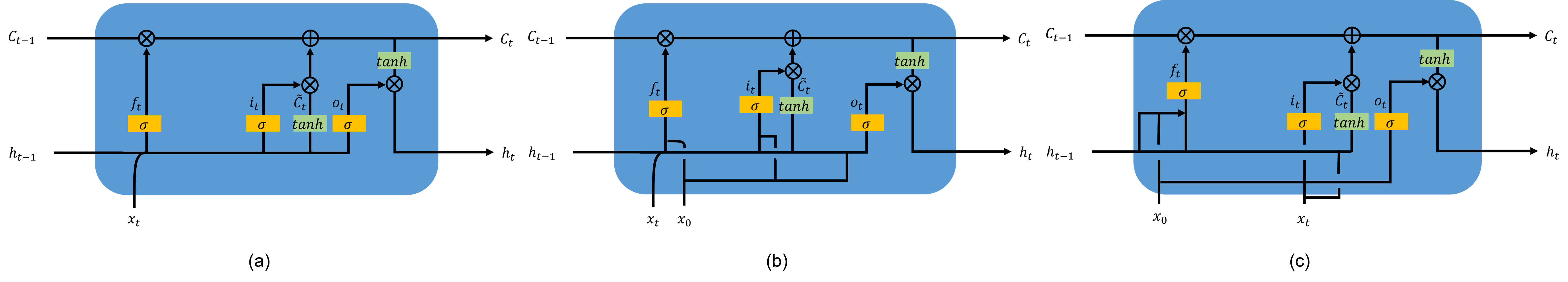}
\caption{Three compared models for the ablation study. (a) original LSTM w/o-$x_0$  that does not contain $x_0$, (a) LSTM w/-$x_0$ in a naïve way that takes $h_{t-1}$, $x_t$, and $x_0$ in a bundle as input (3) LSTM w/-$x_0$ in a sophisticated way that uses $x_0$ instead of $x_t$ or $h_{t-1}$ by considering the role of each gate.}
\label{fig:three model}
\end{figure*}

To demonstrate the importance of our IEU adding a new elevation gate and using $x_0$ appropriately, we compare the evaluation results from four models using four recurrent units, i.e., original LSTM, LSTM taking additional current information in a naïve way, LSTM taking additionally current information in a sophisticated way, and our IEU. Fig. \ref{fig:three model} depicts the three recurrent units and our IEU.
As shown in Fig. \ref{fig:three model}(a), the first unit, LSTM$_{w/o-x_0}$, does not use the current information $x_0$. This recurrent unit is identical to the original LSTM. The second unit, LSTM$_{w/-x_0-bundle}$, takes $h_{t-1}$, $x_t$, and $x_0$ in a bundle as input. The last unit, LSTM$_{w/-x_0-sophisticated}$ uses $x_0$ instead of $x_t$ or $h_{t-1}$ by fully considering the role of each gate.

\begin{table}[tbh!]
\caption{An ablation study when using different types of information sets in each unit.}
\centering
\begin{tabular}{lrr}
\hline
\multicolumn{1}{c}{\multirow{2}{*}{Model}} & \multicolumn{1}{c}{THUMOS-14} & \multicolumn{1}{c}{TVSeries} \\ \cline{2-3} 
\multicolumn{1}{c}{}                       & \multicolumn{1}{c}{mAP (\%)}   & \multicolumn{1}{c}{mcAP (\%)} \\ \hline
LSTM$_{w/o-x_0}$                                      & 58.5                          & 79.9                         \\
LSTM$_{w/-x_0-bundle}$                              & 58.9                          & 80.5                         \\
LSTM$_{w/-x_0-sophisticated}$                      & 59.4                          & 80.7                         \\
Our IEU                                       & 60.4                          & 81.4                         \\ \hline
\end{tabular}
\label{ablation_tab}
\end{table}

\subsubsection{Taking no additional current action feature}
As presented in Table \ref{ablation_tab}, the performance of LSTM$_{w/o-x_0}$ shows the limitation of not using the current information $x_0$ for OAD. The LSTM$_{w/o-x_0}$ achieves the worst performance compared to other three units that take the current information $x_0$ as input. On THUMOS-14 \cite{b22}, LSTM$_{w/o-x_0}$ achieves at least 0.4\% and at most 1.9\% lower performances. On TVSeries \cite{b9}, LSTM$_{w/o-x_0}$ achieves at least 0.6\% and at most 1.5\% lower performances. This means that taking the current information $x_0$ as input is required for temporal modeling for OAD. 

\subsubsection{Taking additional current information}
As presented in Table \ref{ablation_tab}, the second and third units, LSTM$_{w/-x_0-sohpisticated}$ and LSTM$_{w/-x_0-bundle}$, achieve worse performances than our IEU on both THUMOS-14 \cite{b22} and TVSeries \cite{b9}. This means that the newly-added information elevation gate, which is not included in LSTM$_{w/-x_0}$s, effectively compensates for the limitation of the forget gate. In addition, between them, LSTM$_{w/-x_0-sophisticated}$ outperforms LSTM$_{w/-x_0-bundle}$, which means that assigning inputs fed into each gate in an advanced way by considering each gate is a more effective method of temporal modeling for OAD than using inputs in a bundle.

\subsection{Time for Feature Extraction}
\begin{table}[tbh!]
\caption{Feature extraction processing speed.}
\centering
\begin{tabular}{lrrr}
\hline
 & RGB & TSM \cite{b12} & Optical Flow \\ \hline
Speed (fps)       & 266 & 1,162 & 12   \\
\hline
\end{tabular}%
\label{tab4}
\end{table}
For fast OAD, we also extracted visual features using only RGB frames. Table \ref{tab4} presents the three kinds of speed values required for extracting RGB frames, extracting TSM \cite{b12} features, and optical flows. We measured the feature extraction speed in terms of frames per second (FPS), using a Titan XP GPU. Among them, the speed of extracting optical flows is slowest at 12 fps for dense optical flows in OpenCV. In addition, when adopting visual features obtained from optical flows, we require additional time to extract visual features based on optical flows through an AR network, which is not included in Table \ref{tab4}. Owing to the excessive time cost for optical flows, adopting them makes TML methods infeasible for practical services such as video monitoring for specific actions.


\subsection{Trade-off between performance and speed}
 Table \ref{tab5} presents the trade-off between performance improvement and speed degradation for IEU, compared to the original LSTM. Adopting IEU brings a performance improvement of 5.78\% (i.e., 3.3\%p) but a speed degradation of 12.57\%

\begin{table}[tbh!]%
\caption{Trade-off of adopting IEU instead of LSTM between efficiency and effectiveness.}
\centering
\begin{tabular}{lrrr}
\hline
\multicolumn{1}{l}{\quad}    & LSTM & Our IEU & $\Delta$val    \\ \hline
Speed (fps)       & 53,934            & 47,154         & -12.57\%      \\
mAP (\%)   & 57.1             & 60.4          & +5.78\%       \\ \hline
\end{tabular}
\label{tab5}
\end{table}

\subsection{Qualitative Evaluation}
\begin{figure*}[htbp!]
    \centering
    \includegraphics[width=0.82\textwidth]{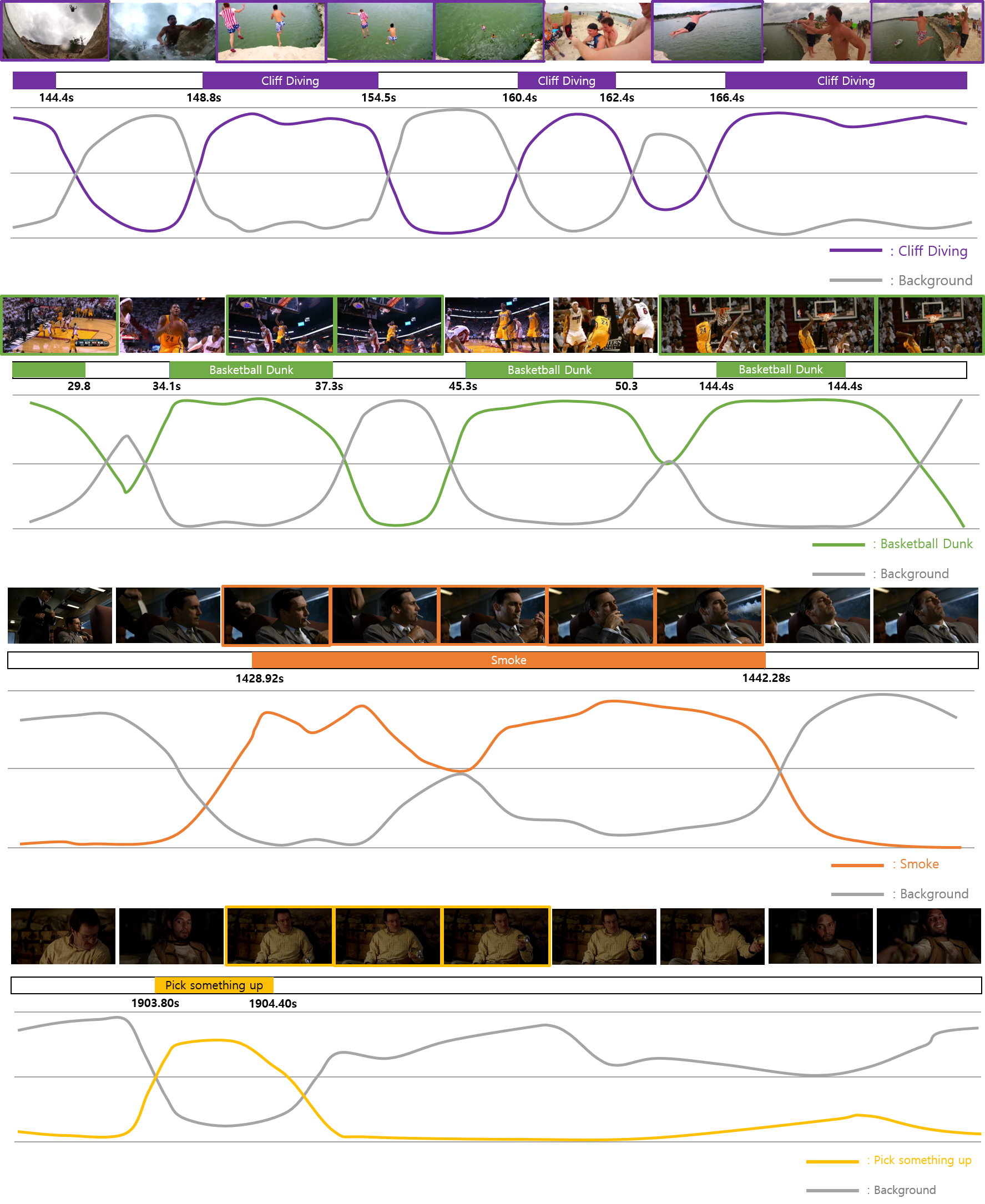}
    \caption{Qualitative evaluation of IEN on THUMOS-14 \cite{b22} and TVSeries \cite{b9}. The frames painted in color represent actions occurring and the graph shown below represents the predicted action probabilities.}
    \label{fig:total_result}
\end{figure*}

As shown in Fig. \ref{fig:total_result}, we visualize qualitative results on THUMOS-14 \cite{b22} and TVSeries \cite{b9}. The top two parts and bottom two parts of Fig. \ref{fig:total_result} show the results of the qualitative evaluation on THUMOS-14 and TVSeries, respectively. In particular, the IEN shows remarkable qualitative results on THUMOS-14. On THUMOS-14, the IEN determines the current action with high-predictive action probabilities for the action. Moreover, on THUMOS-14, the predicted probabilities between actions and background are distinguishable, as shown in Fig. \ref{fig:total_result}.
\section{Conclusion}
In this paper, we propose an IEU that extends LSTM for OAD, by adding a new elevation gate and taking additional current information as input. The newly-added information elevation gate enables the IEU to compensate for the limitation of the forget gate of LSTM, which may lose the accumulated past information related to the current information. In addition, the appropriate use of the current information as the input of each gate enables the IEU to perform effective temporal modeling for OAD, as the IDU \cite{b4} does. To the best of our knowledge, our IEN is the first attempt that considers the computational overhead for the practical use of OAD. For an efficient design of our IEN based on the IEU, we adopted a fast action recognition network, TSM \cite{b14}, as a feature extractor by using only RGB frames excluding optical flows. Although optical flows are widely used for action recognition and detection methods, extracting optical flows requires heavy computation overhead, which is infeasible for practical usage. Therefore, we use visual features extracted by the TSM taking only RGB frames as input. On two OAD benchmark datasets, THUMOS-14 \cite{b22} and TVSeries \cite{b9}, IEN achieves a per-frame mAP of 60.4\% and 81.4\%, respectively. In particular, on THUMOS-14, our IEN outperforms all the state-of-the-art OAD methods taking two-stream features using both RGB frames and optical flows as input. On TVSeries, our IEN outperforms all the state-of-the-art methods taking features using RGB frames and achieves comparable performance compared to the state-of-the-arts taking the two-stream features using both RGB frames and optical flows as input.


\begin{thebibliography}{00}
\bibitem{b1}
marketsandmarkets, “Video Surveillance Market by System, Offering (Hardware (Camera, Storage Device, Monitor), Software (Video Analytics, Video Management System) \& Service (VSaaS)), Vertical (Commercial, Infrastructure, Residential), and Geography – Global Forecast to 2025”, 2020, https://www.marketsandmarkets.com/MarketReports/video-surveillance-market-645.html (accessed {J}une.8, 2021).

\bibitem{b2}
J. Gao, Z. Yang, and R. Nevatia, “RED: Reinforced encoder-decoder networks for action anticipation,” in Proceedings of the British Machine Vision Conference (BMVC), G. B. Tae-Kyun Kim, Stefanos Zafeiriou, and K. Mikolajczyk, Eds. BMVA Press, Sep. 2017, pp.92.1–92.11. 

\bibitem{b3}
M. Xu, M. Gao, Y.-T. Chen, L. S. Davis, and D. J. Crandall, “Temporal recurrent networks for online action detection,” in Proceedings of the IEEE/CVF International Conference on Computer Vision (ICCV), Oct. 2019, pp.5532–5541.

\bibitem{b4}
H. Eun, J. Moon, J. Park, C. Jung, and C. Kim, “Learning to discriminate information for online action detection,” in Proceedings of the IEEE/CVF Conference on Computer Vision and Pattern Recognition, 2020, pp. 809–818.

\bibitem{b5}
H. Eun, J. Moon, J. Park, C. Jung, and C. Kim, "Temporal filtering networks for online action detection," Pattern Recognition, vol. 111, p. 107695, 2021.

\bibitem{b6}
S. Hochreiter and J. Schmidhuber, “Long short-term memory,” Neural  Computation, vol. 9, no. 8, pp. 1735–1780, 1997.

\bibitem{b7}
Z. Shou, D. Wang, and S.-F. Chang, “Temporal action localization in untrimmed videos via multi-stage CNNs,” in Proceedings of the IEEE Conference on Computer Vision and Pattern Recognition, 2016, pp. 1049–1058.

\bibitem{b8}
T. Lin, X. Zhao, H. Su, C. Wang, and M. Yang, “Bsn: Boundary sensitive network for temporal action proposal generation,” in Proceedings of the European Conference on Computer Vision (ECCV), 2018, pp. 3–19.

\bibitem{b9}
R. De Geest, E. Gavves, A. Ghodrati, Z. Li, C. Snoek, and T. Tuytelaars, “Online action detection,” in European Conference on Computer Vision. Springer, 2016, pp. 269–284.

\bibitem{b10}
K. Cho, B. van Merri enboer, C. Gulcehre, D. Bahdanau, F. Bougares, H. Schwenk, and Y. Bengio, “Learning phrase representations using RNN encoder–decoder for statistical machine translation,” in Proceedings of the 2014 Conference on Empirical Methods in Natural Language Processing (EMNLP). Doha, Qatar: Association for Computational Linguistics, Oct. 2014, pp. 1724–1734.

\bibitem{b11}
D. Tran, L. Bourdev, R. Fergus, L. Torresani, and M. Paluri, “Learning spatiotemporal features with 3d convolutional networks,” in Proceedings of the IEEE International Conference on Computer Vision, 2015, pp.4489–4497.

\bibitem{b12}
J. Carreira and A. Zisserman, “Quo vadis, action recognition? a new model and the kinetics dataset,” in Proceedings of the IEEE Conference on Computer Vision and Pattern Recognition, 2017, pp. 6299–6308.

\bibitem{b13}
Y. Wan, Z. Yu, Y. Wang, and X. Li, “Action recognition based on two-stream convolutional networks with long-short-term spatiotemporal features,” IEEE Access, vol. 8, pp. 85 284–85 293, 2020.

\bibitem{b14}
J. Lin, C. Gan, and S. Han, “TSM: Temporal shift module for efficient video understanding,” in Proceedings of the IEEE/CVF International Conference on Computer Vision, 2019, pp. 7083–7093.

\bibitem{b15}
C. Zhang, Y. Zou, G. Chen and L. Gan, “PAN: Towards fast action recognition via learning persistence of appearance,” 2020, arXiv:2008.03462.[Online]. Available:https://arxiv.org/abs/2008.03462

\bibitem{b16}
Karen Simonyan and Andrew Zisserman. "Very deep convolutional networks for large-scale image recognition," arXiv:1409.1556, 2014.

\bibitem{b17}
Yuanjun Xiong, Limin Wang, Zhe Wang, Bowen Zhang, Hang Song, Wei Li, Dahua Lin, Yu Qiao, Luc Van Gool, and Xiaoou Tang. CUHK ETHZ SIAT submission to activitynet challenge 2016. arXiv:1608.00797, 2016.

\bibitem{b18}
H. Xia and Y. Zhan, "A survey on temporal action localization," in IEEE Access, vol. 8, pp. 70477-70487, 2020.

\bibitem{b19}
K. Yun, Y. Kwon, S. Oh, J. Moon, and J. Park, “Vision-based garbage dumping action detection for real-world surveillance platform,” ETRI Journal, vol. 41, no. 4, pp. 494–505, 2019.

\bibitem{b20}
Q. Han, H. Zhao, W. Min, X. Zhou, K. Zuo, and R. Liu, "A two-stream approach to fall detection with mobileVGG," in IEEE Access, vol. 8, pp. 17556-17566, 2020.

\bibitem{b21}
D-A. Clevert, T. Unterthiner, and S. Hochreiter, "Fast and accurate deep network learning by exponential linear units (ELUs),". in Proceedings of the International Conference Learning Representations (ICLR), 2016.

\bibitem{b22}
Y. G. Jiang, J. Liu, A. Roshan Zamir, G. Toderici, I. Laptev, M. Shah, and R. Sukthankar. Thumos challenge: Action recognition with a large number of classes, 2014. http://crcv.ucf.edu/THUMOS14/.

\end{thebibliography}
\end{document}